\documentclass[10pt,twocolumn,letterpaper]{article}

\usepackage[pagenumbers]{iccv} 

\definecolor{iccvblue}{rgb}{0.21,0.49,0.74}
\usepackage[pagebackref,breaklinks,colorlinks,allcolors=iccvblue]{hyperref}
\usepackage{hyperref}       
\usepackage{url}            
\usepackage{booktabs}       
\usepackage{amsfonts}       
\usepackage{nicefrac}       
\usepackage{microtype}      
\usepackage{xcolor}         
\definecolor{aliceblue}{RGB}{204,232,207}
\usepackage{colortbl}
\usepackage{amsmath} 
\usepackage{graphicx}
\usepackage{tikz,graphics,color,float,epsf,caption,subcaption}
\usepackage{booktabs, 
            makecell, multirow, tabularx}
\usepackage{algorithm}
\usepackage{algorithmic}

\usepackage[toc]{appendix}
\usepackage{etoc}
\usepackage{minitoc}

\newcommand{\up}[1]{\textcolor{OliveGreen}{\small \ $\uparrow${#1}}}
\newcommand{\downbad}[1]{\textcolor{Maroon}{\small \ $\downarrow${#1}}}
\newcommand{\down}[1]{\textcolor{OliveGreen}{\small \ $\downarrow${#1}}}
\newcommand{\upbad}[1]{\textcolor{Maroon}{\small \ $\uparrow${#1}}}
\newcommand{\basex}[1]{\textcolor{gray!50}{\small \ $\uparrow${#1}}}

\newcommand{\modelname}{\textsc{AID}\xspace}

\def\paperID{14261} 
\def\confName{ICCV}
\def\confYear{2025}

\title{Attention Hijackers: Detect and Disentangle Attention Hijacking in LVLMs for Hallucination Mitigation}

\author{
Beitao Chen\textsuperscript{1}\\
{\tt\small chenbeitao@gmail.com}
\and
Xinyu Lyu\textsuperscript{2,4}\\
{\tt\small xinyulyu68@gmail.com}
\and
Lianli Gao\textsuperscript{1}
\\
{\tt\small lianli.gao@uestc.edu.cn}
\and
Jingkuan Song\textsuperscript{1,3}\\
{\tt\small jingkuan.song@gmail.com}
\and
Heng Tao Shen\textsuperscript{1,3}\\
{\tt\small shenhengtao@hotmail.com}
\and
\textsuperscript{1}Center for Future Media, University of Electronic Science and Technology of China \\
\textsuperscript{2}Southwestern University of Finance and Economics, Chengdu, China \\
\textsuperscript{3}Tongji University \\
\textsuperscript{4}Engineering Research Center of Intelligent Finance, Ministry of Education \\
}

\begin{document}
\maketitle
\begin{abstract}
Despite their success, Large Vision-Language Models (LVLMs) remain vulnerable to hallucinations.
    While existing studies attribute the cause of hallucinations to insufficient visual attention to image tokens, our findings indicate that hallucinations also arise from interference from instruction tokens during decoding.
    Intuitively, certain instruction tokens continuously distort LVLMs' visual perception during decoding, hijacking their visual attention toward less discriminative visual regions. This distortion prevents them integrating broader contextual information from images, ultimately leading to hallucinations.
    We term this phenomenon “\textbf{Attention Hijacking}”, where disruptive instruction tokens act as “\textbf{Attention Hijackers}”.
    To address this, we propose a novel, training-free strategy namely \textbf{Attention HIjackers Detection and Disentanglement} (\textbf{AID}), designed to isolate the influence of Hijackers, enabling LVLMs to rely on their context-aware intrinsic attention map. 
    Specifically, AID consists of three components: 
    First, \textbf{Attention Hijackers Detection} identifies Attention Hijackers by calculating instruction-driven visual salience.
    Next, \textbf{Attention Disentanglement mechanism} is proposed to mask the visual attention of these identified Hijackers, 
    and thereby mitigate their disruptive influence on subsequent tokens. 
    Finally, \textbf{Re-Disentanglement} recalculates the balance between instruction-driven and image-driven visual salience to avoid over-masking effects.
    Extensive experiments demonstrate that AID significantly reduces hallucination across various LVLMs on several benchmarks. 
    Project page: \url{https://github.com/BT-C/AID}.
\end{abstract}    
\section{Introduction}
\label{sec:intro}

\begin{figure}[t]
  \centering
  \includegraphics[width=1\linewidth]{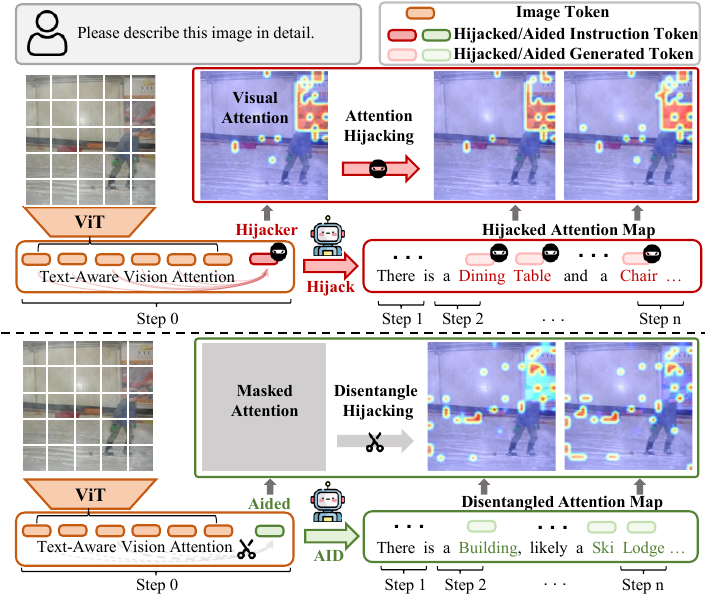}

   \caption{
   \textbf{Upper part:} shows Attention Hijacking, where the Hijacker’s erroneous attention map disrupts generated tokens, causing hallucinations. \textbf{Lower part:} illustrates how our proposed AID isolates Hijacker's influence, allowing the LVLM to rely on its context-aware attention map, thus reducing hallucinations.}
   
   
   \label{fig:abstract}
\end{figure}

    %
    Large Vision-Language Models (LVLMs) have recently demonstrated remarkable advancements, showcasing impressive performance across a wide range of tasks.~\citep{GPT4V, alayrac2022flamingo, li2023blip, zhu2023minigpt, bai2023qwen, dai2023instructblip, wang2023cogvlm, driess2023palm}.
    Despite their remarkable versatility, LVLMs encounter a significant challenge known as hallucination. 
    Specifically, this issue arises when there is a discrepancy between the textual output generated by the model and the actual visual input~\cite{li2023evaluating, gunjal2023detecting, liu2023mitigating, lovenia2023negative}. Such mismatches may manifest as irrelevant or nonsensical responses, or in inaccuracies regarding colors, quantities, and locations of objects that do not appear in the image.


\begin{figure}
  \centering
  \includegraphics[width=1\linewidth]{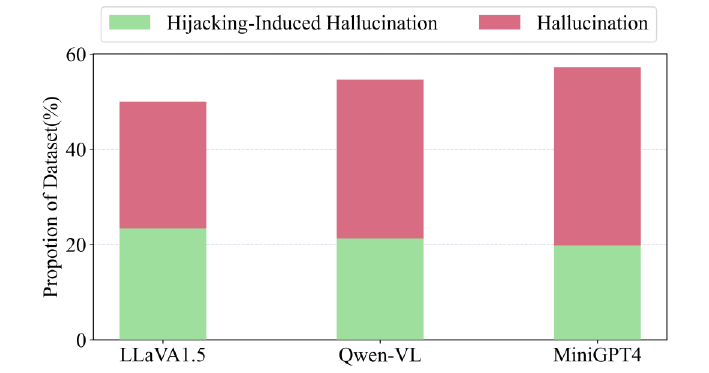}

   \caption{
    \textbf{The proportion of Hijacking-Induced hallucinations} across different LVLMs.
    }
   \label{fig:proportion}
\end{figure}

    Existing research can be categorized into two types. The first relies on supervised fine-tuning~\cite{chen2024alleviating, liu2023mitigating, yu2023rlhf, sun2023aligning,zhao2023beyond}, 
    which requires the creation of high-quality datasets, involving significant training costs. To make hallucination mitigation training-free, recent works~\cite{an2024agla, huo2024self, liu2024paying, gong2024damro,xing2024mitigating} have explored attention mechanisms and proposed inference-based methods to reduce hallucinations. 
    While existing studies attempts to make models more interactive with image tokens, they overlook the visual influence of instruction tokens during LVLMs' decoding process. This oversight limits the effectiveness of these methods in hallucination mitigation.

    
    
    Our observations show that visual information conveyed by instruction tokens can significantly distort LVLMs’ visual perception when generating subsequent tokens, accounting for up to 46.7\% of total hallucinations (Fig.~\ref{fig:proportion}).
    Specifically, the instruction token prompts the LVLMs to repeatedly attend to the same image area with similar attention maps (Fig.~\ref{fig:abstract}).
    The underlying reason is that the attention distribution of LLM decoder on image tokens closely mirrors that of the visual encoder (i.e. ViT~\cite{vit}), with both distributions concentrating on background elements rather than the intended objects in the image (Fig.~\ref{fig:drawback_of_vit}).
    This misdirection causes the model to persistently focus on these misleading regions during subsequent auto-regressive processes, ignoring the broader contextual area and ultimately leading to hallucinations. 
    As illustrated in Fig.~\ref{fig:abstract}, Attention Hijacker makes model continuously focus on negative local regions around the \textit{poster} while ignoring the surrounding context of the \textit{building} and the \textit{skier} during decoding process, resulting in hallucinations such as \textit{dining table} and \textit{chair}.
    We term this pattern as \textbf{Attention Hijacking}, and we refer to the instruction tokens responsible for this pattern as \textbf{Attention Hijackers}.

    The above challenges motivate us to study two problems: (1) how to identify Attention Hijackers among instruction tokens, and (2) how to efficiently isolate the influence of identified hijackers, allowing the model to independently explore appropriate visual interaction patterns.
    Inspired by~\cite{orgad2024llms}, our key intuition is that LVLMs' internal visual attention reflect the likelihood of hallucinations, which can be mitigated by modifying the attention weights of these hijacker tokens during the decoding process. 
    Accordingly, we identify Attention Hijackers by assessing the volume of visual information that instruction tokens impose on subsequently generated tokens via visual attention map. The rationale is that the more visual information an instruction token contributes, the more likely generated token's visual attention may get hijacked by it. Once hijackers are identified, their disruptive influence on less discriminative visual regions can be isolated, restoring the model’s context-aware attention capability. 
    
    Based on the above analysis, we propose a novel, training-free strategy, \textbf{Attention Hijackers Detection and Disentanglement},  
    which isolates the influence of Hijackers, allowing LVLMs to utilize their context-aware intrinsic attention map.
    Specifically, the process begins with \textbf{Attention Hijackers Detection}, which identifies Attention Hijackers by calculating instruction-driven visual salience. Next, \textbf{Attention Disentanglement} mechanism is introduced to mask the visual attention of the previously identified hijackers. Finally, to prevent excessive masking of the visual perception of non-hijacker instruction tokens made by Attention Disentanglement, \textbf{Re-Disentanglement} is proposed to adjust the balance between instruction-driven and image-driven visual salience, ensuring effective disentanglement.

    In conclusion, our main contributions are summarized as follows:
    \begin{enumerate}
        \item We, for the first time,  identify that hallucinations stem from the disruptive interactions with certain instruction tokens (i.e., Attention Hijacker) and conduct an in-depth analysis of how these hijackers interfere model's visual attention, elucidating the mechanisms by which this disruption leads to hallucinations.
        \item Based on our exploration, we propose \textbf{A}ttention H\textbf{I}jackers Detection and \textbf{D}isentanglement (\textbf{AID}), a training-free strategy that first locates potential Attention Hijackers and disentangles hijacked attention from the Attention Hijacker by cutting-off the visual source associated with the Attention Hijacker.
        \item Through comprehensive experiments, we demonstrate our proposed AID achieves significant improvements in alleviating hallucinations without requiring additional training or the use of external tools.

    \end{enumerate}

\section{Related Work}
\label{sec:rel}

\noindent \textbf{Hallucination in LVLMs.} 
    In Vision-Language Models (VLMs), ``object hallucination"—where models generate plausible but mismatched or missing objects in images—is a known issue~\cite{rohrbach2018object,biten2022let,li2023evaluating}. Traditional VLMs address this through fine-grained contrastive learning~\cite{zeng2021multi}, ROI feature fusion ~\cite{biten2022let}, and data augmentation to reduce co-occurrence patterns~\cite{kim2023exposing}. However, these techniques to autoregressive LVLMs is challenging~\cite{kaplan2020scaling,wei2022emergent}. 
   Recent research on LVLMs focuses on developing evaluation and detection methods \cite{wang2023evaluation,liu2023aligning,li2023evaluating,lovenia2023negative} for hallucination, such as the CHAIR metric\cite{rohrbach2018object} for caption accuracy and POPE\cite{li2023evaluating} for binary object recognition. Further progress includes creating refined datasets for fine-tuning 
 \cite{gunjal2023detecting,li2023m,liu2023aligning}, training post-hoc revisors \cite{zhou2023analyzing}, and using factually augmented Reinforcement Learning from Human Feedback (RLHF) \cite{sun2023aligning}.

    \noindent \textbf{Attention Deficits of LVLMs.}
    A key approach to reduce hallucinations in LVLMs is to address attention deficits by adjusting or decoding strategies~\cite{an2024agla, huo2024self, liu2024paying, gong2024damro,xing2024mitigating}. Early studies~\cite{darcet2023vision, gong2024damro} showed that LVLMs often focus on global image features, neglecting prompt-relevant details~\cite{an2024agla}, a limitation linked to the Vision Transformer encoder~\cite{alexey2020image}. To counter this, some methods adaptively amplify attention weights for relevant image tokens~\cite{liu2024paying}, while others filter effective image information based on attention scores, reducing hallucinations via contrastive decoding~\cite{huo2024self}. Recent advances, like Concentric Causal Attention, further mitigate object hallucinations by addressing long-term decay in Rotary Position Encoding~\cite{xing2024mitigating}.


Existing methods attempt to mitigate hallucinations by adjusting LVLMs' attention to relevant visual elements, yet they often overlook disruptions caused by specific instruction tokens. Our approach isolates disruptive influences, restoring accurate attention and significantly reducing hallucinations across benchmarks.

\section{Method}
\label{sec:Method}

 \begin{figure}
  \centering
  \includegraphics[width=1\linewidth]{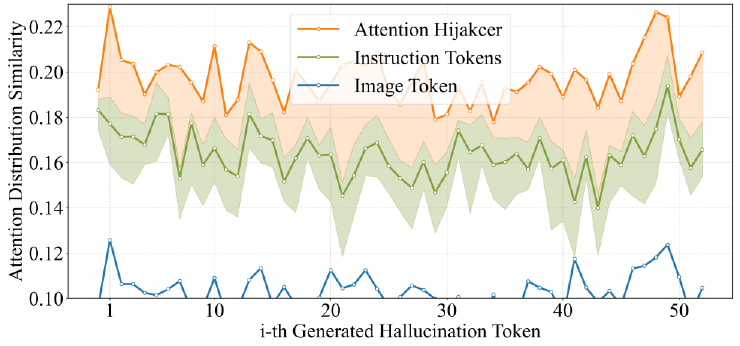}
  
   \caption{
     \textbf{Illustration of Attention Hijacking.} The Attention Hijacker (\textcolor[HTML]{EF9D55}{orange} curve) shows a visual attention distribution more aligned with the hallucination token than with the image token (\textcolor[HTML]{6B9AB6}{blue} curve), indicated by consistently higher similarity values.
    }
   \label{fig:last_impact}
\end{figure}

\subsection{Problem Formulation}
\label{sec:problem_formulation}

    \noindent \textbf{Autoregressive language decoders.} 
    %
    %
    Each attention head in a single layer performs repeated attention operations with identical input shapes:
    \begin{equation}
    \label{eq:qkv}
    O_h = A_h V_h, \quad A_h = softmax \left(\frac{Q_h K_h^{\top}}{\sqrt{d_k}}\right).
    \end{equation}
    Each attention head $h$ uses its own set of queries $ Q_h \in \mathbb{R}^{n \times d_k}$, keys $ K_h \in \mathbb{R}^{n \times d_k}$, and values $ V_h \in \mathbb{R}^{n \times d_k}$, where $n$ is the sequence length and $d_k$ represents hidden dimensions.
    The output $ O_h \in \mathbb{R}^{n \times d_k}$ is calculated by multiplying $ V_h$ with attention weights $ A_h \in \mathbb{R}^{n \times n}$, with each row representing the weights for each token during feature mixing.
    %

\begin{figure}[t]
  \centering
  \includegraphics[width=1\linewidth]{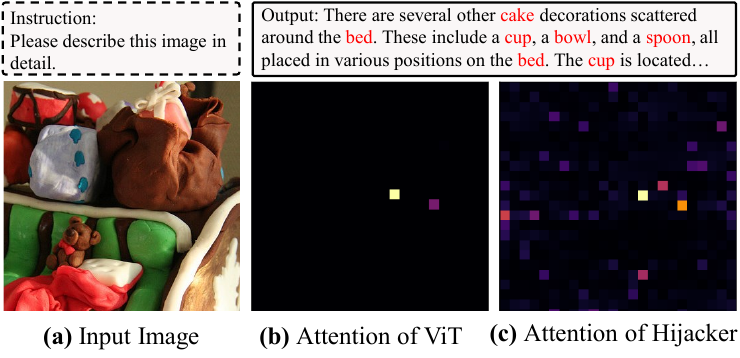}
   \vspace{-5mm}
   \caption{
  \textbf{Attention deficiencies of Visual Encoder (ViT).} Vision Transformer overemphasis background regions with high-norm outlier tokens, resulting in hallucinations though Attention Hijacking.  
    }
   \label{fig:drawback_of_vit}
\end{figure}

\subsection{Analysis of Attention Hijacking}
\label{sec:Attention_Hijacking_Induce_Hallucinations}





    \noindent \textbf{Observations of Attention Hijacking.} 
    (i) To examine how attention hijacking occurs, we measured the similarity in visual attention between certain instruction tokens and hallucinated tokens, as shown in Fig.~\ref{fig:last_impact}. The results reveal that hallucinated tokens get persistently interfered by the Attention Hijacker (\textcolor[HTML]{EF9D55}{orange} curve), maintaining a high Attention Distribution Similarity scores across all hallucination tokens. 
    (ii) Furthermore, we observed that Attention Hijacker (\textcolor[HTML]{EF9D55}{orange} curve) exhibits a higher Attention Distribution Similarity than the image token (\textcolor[HTML]{6B9AB6}{blue} curve), suggesting that generated tokens are more affected by certain instruction tokens (i.e., Attention Hijackers) than by actual image content.
    (iii) Lastly, we observed that different instruction tokens exert varying degrees of influence on the model, as indicated by different levels of Attention Distribution Similarity (\textcolor[HTML]{91A34A}{green} \textit{v.s.,} \textcolor[HTML]{EF9D55}{orange}) in Fig.~\ref{fig:last_impact}. This motivates us to develop a hijacker detection mechanism to identify tokens with the strongest interference during decoding process, as elaborated in Sec.~\ref{sec:Hijackers_Detection}.

\begin{figure*}
  \centering
  \includegraphics[width=1\linewidth]{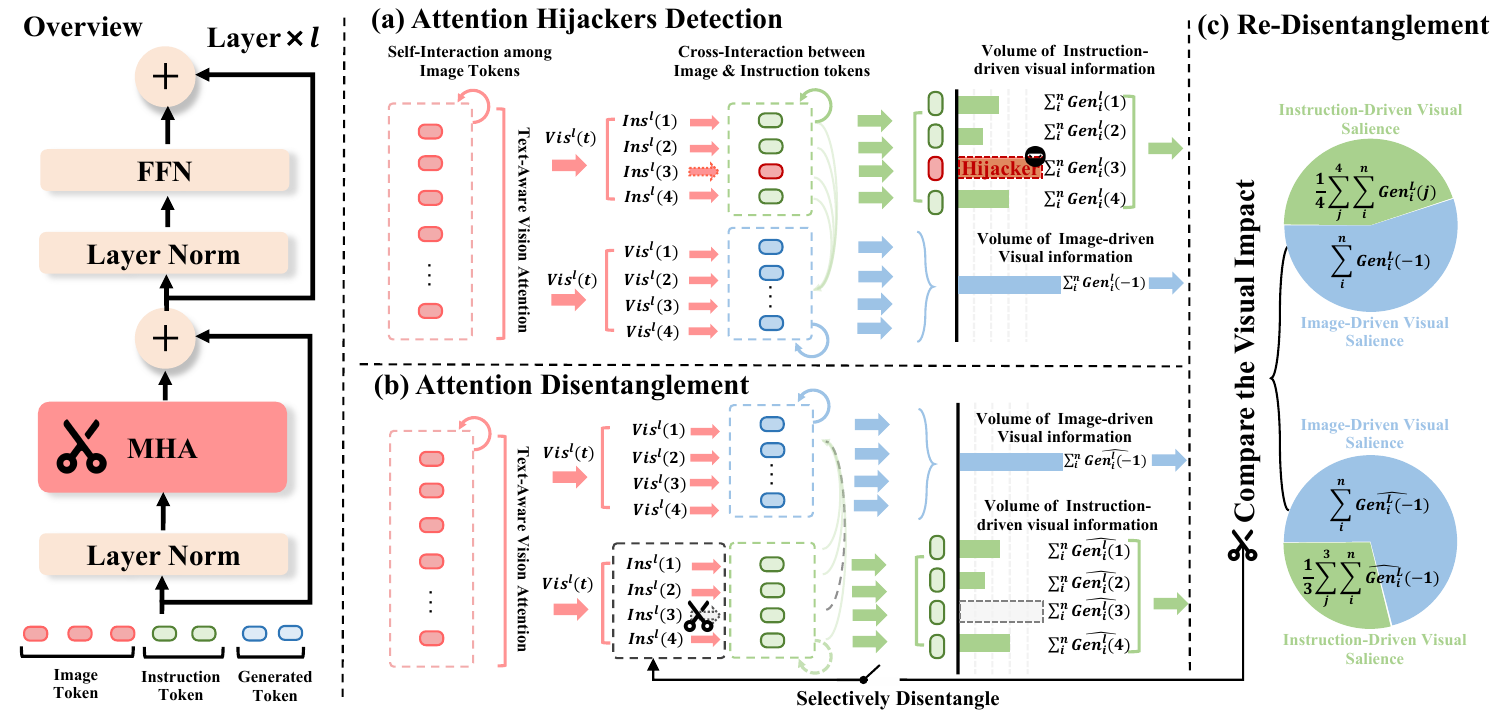}

   \caption{
   \textbf{The framework of our AID.} The proposed framework consists of three key components: Attention Hijackers Detection, which identifies hijackers by calculating instruction-driven visual salience; Attention Disentanglement, which masks the visual attention of identified hijackers to isolates the influence of Hijackers; and Re-Disentanglement, which adjusts the balance between instruction-driven and image-driven salience to prevent over-masking and ensure effective Disentanglement.}
   \label{fig:framework}
\end{figure*}

    \noindent \textbf{Mechanism of Attention Hijacking's Sustainable Impact.} 
    In autoregressive process, generated tokens are influenced by preceding instruction tokens, with this impact reinforced by the KV-cache~\cite{kvchae}, which retains and reuses past attention patterns. Certain instruction tokens, or ``Attention Hijackers," impose strong, localized attention that persists through the KV-cache, causing subsequent tokens to focus on irrelevant visual cues rather than adapting to broader image contexts. This KV-cache persistence amplifies hijacker influence, directing attention toward misleading regions.

    \noindent \textbf{Why Attention Hijacking Leads to Hallucinations?}  
    We hypothesize that the Vision Transformer (ViT) overemphasizes background regions with high-norm outlier tokens, resulting in hallucinations though Attention Hijacking. This misdirects the model’s focus to irrelevant areas, causing it to overlook essential details and misinterpret the visual context.
    To verify this hypothesis, we visualize the visual attention of the Vision Transformer (ViT) and that of the Attention Hijacker in Fig.~\ref{fig:drawback_of_vit}. Fig.~\ref{fig:drawback_of_vit} (b) illustrates the influence of high-norm outlier tokens on ViT's visual attention, which typically correspond to background areas associated with negative visual priors~\cite{darcet2023vision}. Meanwhile, the attention of the instruction token depicted in Fig.~\ref{fig:drawback_of_vit} (c) exhibits a distribution similar to ViT's in Fig.~\ref{fig:drawback_of_vit} (b). This similarity may arise because, when these image tokens are projected into the LLM, their high attention values in the visual encoder lead the model to disproportionately focus on them, resulting in the neglect of local details in other patches. This finding aligns with the conclusion elaborated in~\cite{gong2024damro}.
    
    \noindent \textbf{Solution.} The limitations of Vision Transformers (ViT) can be addressed by continuously increasing the visual perception and fine-grained understanding of the model.
    However, the generation of fine-grained annotated data and the training of visual encoders are both prohibitively expensive~\cite{chen2023double_vision_tower}. 
    Inspired by~\cite{zhao2024first}, our core insight is that LVLMs' internal visual attention reflects the likelihood of hallucinations, which can be reduced by adjusting the attention weights of hijacker tokens during decoding. To achieve this, we propose Attention Hijackers Detection and Disentanglement (AID) (shown in Fig.~\ref{fig:framework}) to firstly identify Attention Hijackers by analyzing the visual influence that instruction tokens impose on subsequent tokens through attention analysis. Then, by adaptively masking the attention maps of these hijackers, we redirect the model’s focus from less informative visual regions to broader contextual areas in the image. The process for identifying Attention Hijackers and restoring the model's context-aware visual attention are discussed in the following sections.

\subsection{Attention Hijackers Detection}
\label{sec:Hijackers_Detection}


     Based on findings from Sect.~\ref{sec:Attention_Hijacking_Induce_Hallucinations}, we propose the Attention Hijackers Detection method to detect Attention Hijackers by calculating the volume of instruction-driven visual information across all tokens, as shown in Fig.~\ref{fig:framework} (b).
     This process involves two integration phases, i.e., Self-Interaction of Image Tokens and Cross-Interaction between Image \& Instruction tokens, followed by a attention-based information aggregation phase, i.e., Visual Information Transmission to Generated Tokens.    
    \noindent \textbf{Self-Interaction among Image Tokens.} 
    The self-interaction of image tokens redistributes model's visual attention, making later tokens carry more than earlier ones (Fig. \ref{fig:ablation_area_mask}) and providing varying levels of visual information to the hijacker. Therefore, prioritizing these calculations is essential. The self-interaction among image tokens is calculated as follows:
    \begin{equation}
    \label{eq:vis_aggregation}
    	Vis^{l}(t) = \begin{cases}
        1, & l = 0 \\ 
    	\sum^{j \in \mathbb{I}_{(t,vis)}^l} (1 + w_j) Vis^{l - 1}(j), & l > 0
    	\end{cases},
    \end{equation}
    where $Vis^l(t)$ represents visual information of $t\,th$ image token on $l\,th$ layer, $\mathbb{I}_{(t, vis)}^l$ indicates index of visual attention weight associated with the $t\,th$ image token's self-interaction and $w_j$ is the $j\,th$ visual attention weight. We initialize visual information in $0\,th$ layer to be identical across all image tokens, allowing it to evolve through subsequent interactions.
    
    \noindent \textbf{Cross-Interaction between Image \& Instruction tokens.} 
    In this step, the visual information get integrated into the instruction tokens, either directly from the image or indirectly from the previous instruction tokens. The calculations are shown below:
    \begin{equation}
    \label{eq:instruction_aggregation}
    	Ins^{l}(t) = \begin{cases}
        0, & l = 0 \\ 
    	\sum^{j \in \mathbb{I}_{(t,vis)}^l} w_j Vis^{l - 1}(j)  \\   
        +\sum^{j \in \mathbb{I}_{(t, ins)}} (1 + w_j) Ins^{l - 1}(j), & l > 0
    	\end{cases},
    \end{equation}
    where $Ins^l(t)$ denotes the visual information of $t$-th instruction token on $l$-th layer, with $\mathbb{I}_{(t, ins)}^l$ indicating the index of visual attention weight associated with the $t$-th instruction token and $w_j$ as the $j$-th visual attention weight. In $0\,th$ layer, visual information of all instruction tokens is initialized to zero, as these tokens originate from user prompts and lack visual modality data. In subsequent layers, each instruction token integrates visual information both directly from image tokens and indirectly from previous-layer instruction tokens that have absorbed visual information.
    
    \noindent \textbf{Visual Information Transmission to Generated Tokens.} 
    This step measures the volume of visual information from each instruction token to generated tokens, which primarily derives from three components:
    (1) the instruction token, and (2) the tokens generated in previous step. The calculations are shown below:
\begin{small} 
    \begin{equation}
    \label{eq:gen_token_aggregation}
    	Gen_{i}^{l}(t) = \begin{cases}
        w_t Ins^{l}(t) \\ 
        + \sum^{j \in \mathbb{I}_{(t,Gen)}^l} w_j Gen_{j}^{l}(t), & l = 0, t \neq -1 \\ 
    	w_t Ins^{l}(t) + (1 + w_i)Gen_{i}^{l-1}(t)\\
        + \sum^{j \in \mathbb{I}_{(t,Gen)}^l} w_j Gen_{j}^{l}(t), & l > 0, t \neq -1  \\   
    	\end{cases},
    \end{equation}    
\end{small}where $Gen_{i}^{l}(t)$ denotes the visual information contributed by the $t$-th instruction token to the $i$-th generated token. $\mathbb{I}_{(i,vis)}^l$ represents the index of visual attention weight associated with the $i$-th generated token. 
    Finally, find $\mathbb{S}_{H} = \{t_1, t_2, ..., t_j \}$ such that $\left| \mathbb{S}_{H} \right| = k $ and $\forall x \in \mathbb{S}_{H}$, $\forall y \not\in \mathbb{S}_{H}$, where $\mathbb{S}_{H}$ is the \textbf{Attention Hijackers}.
    \begin{equation}
    \label{eq:hijacker_detection}
        \sum^{i \in I_{gen}} Gen_{i}(x) > \sum^{i \in I_{gen}} Gen_{i}(y).
    \end{equation}

\subsection{Attention Disentanglement}
\label{sec:Attention_Disengagement}
    After identifying Attention Hijackers in Sect.~\ref{sec:Hijackers_Detection},
    we implement the Attention Disentanglement mechanism to disentangle model's visual attention from hijacker tokens. This method selectively blocks hijackers from receiving visual information while preserving relevant visual cues for non-hijacking tokens. Thereby, it refocuses the model on accurate visual regions, reducing hallucinations and enhancing the alignment with intended visual content.

   Specifically, we disconnect all visual information received by the hijacker during the visual-textual interaction, as shown in Fig.~\ref{fig:framework}(b). 
   Practically, we cut-off all tokens carrying image information (including image tokens and non-hijacking instruction tokens with visual information) from interacting with the hijacker, while ensuring that non-hijacking instruction tokens still receive image information as specified in Eqn.~\ref{eq:instruction_aggregation}. The specific cut-off operation is shown in the following formula:
    \begin{equation}
    \label{eq:Attention_Disengagement}
    	DIns^{l}(t) = \begin{cases}
        
        
        Ins^{0}(t), & l = 0 \\ 
        
        
        
        \sum^{j \in \mathbb{I}_{(t, ins)}} (1 + w_j) DIns^{l - 1}(j), & l \leq l_k
    	\end{cases},
    \end{equation}
    where $DIns^{l}(t)$ represents the intermediate instruction token generated after Attention Disentanglement, containing only results related to the hijacker, as shown in:
    \begin{equation}
    \label{eq:combine_Attention_Disengagement}
    	Ins^{l}(t) =  DIns^{l}(t) , t \in \mathbb{S}_{H}.
    \end{equation}
    For other non-hijacking instruction tokens, the standard interactions defined in Eqn.\ref{eq:instruction_aggregation} are retained. 

\begin{table*}[htbp]
    \centering
            \renewcommand{\arraystretch}{1.05}
    \tabcolsep=0.25cm
    \small
      \begin{tabular}{l|ccc|ccc|ccc}
      \toprule
      \multirow{2}[2]{*}{\textbf{Methods}} & \multicolumn{3}{c|}{\textbf{LLaVA-1.5}} & \multicolumn{3}{c|}{\textbf{MiniGPT-4}} & \multicolumn{3}{c}{\textbf{mPLUG-Owl2}} \\
            & \textit{Random} & \textit{Popular} & \textit{Adversarial} & \textit{Random} & \textit{Popular} & \textit{Adversarial} & \textit{Random} & \textit{Popular} & \textit{Adversarial} \\
      \midrule
      Greedy      & 81.75  & 78.48  & 76.54  & 69.96  & 66.77  & 67.93  & 82.20 & 80.49  & 78.54   \\
      Beam Search & 84.84  & 83.87  & 81.61  & 69.64  & 67.01  & 68.30  & 86.83 & 83.92  & 81.73   \\
      OPERA       & 82.03  & 81.49  & 79.74  & 69.62  & 67.23  & 68.29  & 87.29 & 82.53  & 80.25   \\
      VCD         & 80.42  & 76.05  & 75.95  & 68.82  & 66.80  & 66.26  & 80.50 & 78.51  & 78.67   \\
      DoLa        & 83.56  & 83.37  & 80.97  & 69.85  & 67.70  & 68.65  & 88.11 & 84.75  & 82.17   \\
      \midrule
      \rowcolor{aliceblue} Ours  & \textbf{90.12 } & \textbf{88.77 } & \textbf{85.57 } & \textbf{75.41 } & \textbf{70.01 } & \textbf{70.29} & \textbf{90.81 } & \textbf{86.72 } & \textbf{84.52 } \\
      \bottomrule
      \end{tabular}%
    \caption{Comparison of the average F1-score evaluation results under different settings (i.e., \textit{ Random, Popular, Adversarial}) using various baselines and our AID on offline POPE benchmark~\cite{li2023evaluating, chen2024halc} over five random runs, where higher F1-score indicates better performance. $\dagger$ denote results reproduced with the authors’ codes.
    More detailed statistical results are shown in Tab.1 of Appendix A. }
    \label{tab:res_pope}%
\end{table*}%

\begin{table*}[htbp]
    \centering
        \renewcommand{\arraystretch}{1.0}
    \tabcolsep=0.25cm
    \small
      \begin{tabular}{l|ccc|ccc|ccc}
      \toprule
       \multirow{2}[2]{*}{\textbf{Methods}} & \multicolumn{3}{c|}{\textbf{LLaVA-1.5}} & \multicolumn{3}{c|}{\textbf{MiniGPT-4}} & \multicolumn{3}{c}{\textbf{mPLUG-Owl2}} \\
            & CHAIR$_s\downarrow$ & CHAIR$_i\downarrow$ & Recall$\uparrow$   & CHAIR$_s\downarrow$ & CHAIR$_i\downarrow$ & Recall$\uparrow$   & CHAIR$_s\downarrow$ & CHAIR$_i\downarrow$ & Recall$\uparrow$ \\
      \midrule
            Greedy      & 48.0  & 14.7  & 76.9  & 39.2  & 15.6  & 56.8  & 55.0  & 17.8  & 72.0 \\
            Beam Search & 50.0  & 15.5  & 77.8  & 40.2  & 16.1  & 57.3  & 59.0  & 17.8  & 72.8 \\
            OPERA       & 47.3  & 14.8  & 76.8  & 41.4  & 15.0  & 58.0  & 56.3  & 18.1  & 71.2 \\
            VCD         & 50.4  & 15.7  & 76.4  & 39.6  & 16.2  & 55.5  & 62.8  & 19.8  & 70.7 \\
            DoLa        & 48.6  & 14.6  & 77.5  & 40.0  & 16.3  & 57.5  & 54.8  & 17.4  & 72.7 \\
            \midrule
            \rowcolor{aliceblue} Ours  & \textbf{41.8} & \textbf{13.0 } & 77.1  & \textbf{34.0}  & \textbf{11.0}  & 55.1  & \textbf{53.0}  & \textbf{17.1}  & 71.5  \\
      \bottomrule
      \end{tabular}%
    \caption{Comparison of the average CHAIR evaluation results (instance levels CHAIR$_i$ and sentence levels CHAIR$_s$) and Recall during decoding with different baselines on MSCOCO datasets of five random runs. Experimental results on more backbones (including GLM-4V and Qwen-VL) are shown in Tab.2 of Appendix A.}
    \label{tab:chair}%
    \vspace{-0.5em}
\end{table*}%

\subsection{Re-Disentanglement}
\label{sec:Revise_Disengagement}
    The methods in Sect.~\ref{sec:Attention_Disengagement} risk disrupting valuable visual interactions by broadly masking instruction tokens, which may unintentionally diminish the contributions of non-hijacking instruction tokens, especially when they provide essential contextual information for comprehensive visual understanding.
    
    To address this, we propose the Re-Disentanglement mechanism to selectively disengages the Hijacker's influence by comparing the instruction-driven and image-driven visual salience, as shown in Fig.~\ref{fig:framework} (c).
    Specifically, it compares the visual impact (i.e., instruction-driven visual salience) of each instruction token with that of the image token (i.e., image-driven visual salience), disengaging attention only when the instruction token's impact decreases relative to the image token. A decrease in impact indicates that the instruction token no longer contributes relevant visual context. If no decrease is detected, it suggests that the instruction token is still aligned with the image information, and its visual influence is preserved. This approach ensures that only instruction tokens introducing misdirecting attention are modified, while those providing meaningful visual guidance remain effective in directing the model's focus.
    
     Piratically, we first extend Eqn.~\ref{eq:gen_token_aggregation} to compute the visual information contributed by the image token to the generated token:
    \begin{equation}
    \label{eq:gen_token_aggregation_vis}
    	Gen_{i}^{l}(t) = \begin{cases}
        \sum^{j \in \mathbb{I}_{(i,vis)}^l} w_j Vis^{l}(j), & l = 0, t = -1 \\
        \sum^{j \in \mathbb{I}_{(i,vis)}^l} w_j Vis^{l}(j) \\ 
        + (1 + w_i)Gen_{i}^{l-1}(t), & l > 0, t = -1
    	\end{cases},
    \end{equation}
    where $Gen_{i}^{l}(-1)$ represent the visual information directly from image token. 
    
    Then, the difference between instruction-driven and image-driven visual salience are formulated as follows:
      \begin{small}  
    \begin{equation}
    \label{eq:Filter}
        \sum^{i \in I_{gen}}\left(\sum^{t \in I_{ins}} \frac{Gen_i^{L}(t)}{Gen_i^{L}(-1)}\right) - \sum^{i \in I_{gen}} \left(\sum^{t \in I_{ins}} \frac{\hat{Gen_i^{L}}(t)}{\hat{Gen_i^{L}}(-1)} \right) > 0,
    \end{equation}
    \end{small}where $\hat{Gen_i^{L}}$ denotes the visual information contributed by the $t$-th instruction token to the $i$-th generated token. And, the Hijacker's visual influence is removed once the conditions specified in Eqn.\ref{eq:Filter} are satisfied.

\section{Experiments}
\label{sec:exp}

\begin{table*}[htbp]
    \centering
        \renewcommand{\arraystretch}{1.05}
    \tabcolsep=0.15cm
    \small    
      \begin{tabular}{l|cccc|cccc|cccc}
      \toprule
      \multirow{3}[4]{*}{\textbf{Methods}} & \multicolumn{4}{c|}{\textbf{LLaVA-1.5}} & \multicolumn{4}{c|}{\textbf{MiniGPT-4}} & \multicolumn{4}{c}{\textbf{mPLUG-Owl2}} \\
  \cmidrule{2-13}          & \multicolumn{2}{c}{\textbf{Object-level}$\uparrow$ } & \multicolumn{2}{c|}{\textbf{Attribute-level}$\uparrow$ } & \multicolumn{2}{c}{\textbf{Object-level}$\uparrow$ } & \multicolumn{2}{c|}{\textbf{Attribute-level}$\uparrow$ } & \multicolumn{2}{c}{\textbf{Object-level}$\uparrow$ } & \multicolumn{2}{c}{\textbf{Attribute-level}$\uparrow$ } \\
            & Existence & Count & Position & Color & Existence & Count & Position & Color & Existence & Count & Position & Color \\
      \midrule
      Greedy   & 165.67 & 120.00 & 110.67 & 148.33 & 137.00 & 93.00 & 75.00 & 125.00 & 167.00 & 120.00 & 105.00 & 145.00 \\
      DoLa     & 170.00 & 120.00 & 106.67 & 150.67 & 137.00 & 90.00 & 75.33 & 122.67 & 167.00 & 125.00 & 110.00 & 147.67 \\
      OPERA    & 165.00 & 115.67 & 104.00 & 145.00 & 140.67 & 92.33 & 73.00 & 125.00 & 167.00 & 122.33 & 100.00 & 145.00 \\
      VCD      & 175.33 & 130.33 & 115.00 & 155.00 & 142.00 & 95.33 & 71.33 & 129.00 & 171.33 & 125.00 & 107.33 & 150.00 \\
      HALC     & 167.67 & 121.33 & 106.67 & 150.67 & 140.00 & 92.67 & 71.33 & 122.67 & 167.00 & 120.33 & 108.67 & 145.00 \\
      VASparse & 180.00 & 132.67 & 121.33 & 160.00 & 147.33 & \textbf{98.67} & 78.67 & 133.00 & 175.00 & 130.00 & \textbf{110.67} & 155.00 \\
      \midrule
      \rowcolor{aliceblue} Ours & \textbf{195.00} & \textbf{150.00} & \textbf{130.00} & \textbf{170.00} & \textbf{170.00}  & 96.66 & \textbf{93.33} & \textbf{133.33} & \textbf{180.00} & \textbf{143.33} & 106.66 & \textbf{155.00} \\
      \bottomrule
      \end{tabular}%
    \caption{Results on the subset of the MME benchmark for evaluating object-level and attribute-level VH, where the best performances within each setting are bolded. We randomly run it five times to obtain the average result, with the whole statistical results in Appendix.
    }
    \label{tab:res_mme_hallucination}%
  \end{table*}%


\begin{table*}[t!]
    \centering
    \small
    \resizebox{1.0 \linewidth}{!}{
    \begin{tabular}{lccccccccccc}
    \toprule
    \bf {Method} & \bf gn\_kw\_rec & \bf rec & \bf ocr\_sp & \bf ocr  & \bf ocr\_sp\_rec & \bf ocr\_kw\_rec  & \bf ocr\_gn\_sp & \bf Total  \\
    \midrule
     LLaVA1.5-7B                    & 18.1 \basex{0.0} & 67.6 \basex{0.0} & 17.7 \basex{0.0} & 48.3 \basex{0.0} & \bf60.0 \basex{0.0} & 21.2 \basex{0.0}  & 10.0 \basex{0.0} & 31.1 \basex{0.0} \\
     + VCD                          & 19.2 \up{1.1} & 62.2 \downbad{5.4} & 15.8 \downbad{1.9} &  29.2 \downbad{20.0} & 42.5 \downbad{17.5} & 17.5 \downbad{3.7} & \bf 60.0 \up{50.0} & 30.2 \downbad{1.1} \\ 
     + OPERA                        & \bf 21.8 \up{3.7} & 61.9 \downbad{5.7} & \bf 21.5 \up{3.8} & 51.7 \up{3.4} & 56.2 \downbad{3.8} & 11.2 \downbad{10.0} & 30.0 \up{1.4} & \bf 32.0 \up{0.9} \\ \midrule
    \rowcolor{aliceblue} \bf + Ours &  19.5 \up{1.4} & \bf 70.3 \up{2.7} & 18.5 \up{0.8} & \bf 55.0 \up{6.7} &  58.3 \downbad{1.7} & \bf27.8 \up{6.6} & 12.5 \up{2.5} & 31.8 \up{0.7} \\ 
    \bottomrule
    \end{tabular}}
    \vspace{-1em}
    \caption{\textbf{The MM-Vet evaluation} results encompass multiple complex multimodal tasks. \textit{gn} represents language generation, \textit{kw} indicates knowledge, \textit{sp} denotes spatial awareness, and \textit{rec} stands for recognition.}
    \label{tab:mmvnetsub}
    \vspace{-1em}
\end{table*}

\subsection{Experiment Setup}
\noindent \textbf{Benchmarks.} 
Following common settings~\cite{li2023evaluating, chen2024halc}, 
We evaluate the effectiveness of our AID in VH mitigation across five widely used benchmarks:
(1) the offline Polling-based Object Probing Evaluation (POPE)~\cite{li2023evaluating,chen2024halc} on the MSCOCO dataset;
(2) the CHAIR~\cite{rohrbach2018object} quantitative metrics on MSCOCO dataset~\cite{lin2014microsoft};
(3) three general-purpose benchmarks: LVLM Comprehensive Evaluation (MME)~\citep{fu2023mme}, Multimodal Benchmark (MMBench)~\citep{liu2023mmbench}, and Multimodal Veterinarian (MM-Vet)~\citep{2024MMVet}.


    
\noindent \textbf{Baselines.} 
    We compare our \modelname with greedy decoding, beam search decoding, and several state-of-the-art (SOTA) decoding methods as baselines,
    including DoLa~\cite{chuang2023dola}, OPERA~\cite{huang2023opera}, and VCD~\cite{leng2023mitigating_vcd}. All these methods adopt the beam search decoding strategy with a beam size of 3.

\noindent \textbf{Backbones.} 
    Following previous studies~\cite{li2023evaluating, chen2024halc}, 
    we select five widely used LVLMs families, e.g., LLaVA-1.5~\cite{liu2023visual_llava}, MiniGPT-4~\cite{zhu2023minigpt4}, mPLUG-Owl2~\cite{ye2023mplugowl}, GLM-4V~\cite{glm2024chatglm} and Qwen-VL-Chat~\cite{Qwen-VL}, as the base models for all baselines. We analyze the VH of these LVLMs under different decoding to evaluate the effectiveness of our \modelname.

\noindent \textbf{Settings.} 
    Following HALC~\cite{chen2024halc}, we implement the proposed \modelname using Hugging Face Transformers library~\cite{Wolf2019HuggingFacesTS} and employ beam search for decoding. The LLaVA1.5-7B results are based on version 1.2.2 from the official benchmark repository~\cite{llava:benchmark}. The effectiveness of methods such as VCD, DOLA, and OPERA is evaluated using the HALC implementation~\cite{chen2024halc}. We conduct experiments with a maximum generation length $L_{max}$ of 64 for offline POPE and 1024 for CHAIR. The beam size is set to 3, except for $L_{max}=512$, where it is set to 2. All experiments, including the decoding process of LVLMs, are conducted on eight V100 GPUs. Other methods follow the settings specified in their respective original papers. Further details and results are provided in the Appendix B.

\subsection{Main Results} 
\noindent\textbf{POPE Evaluation.}~
    To evaluate \modelname's capability on object hallucination, following HALC~\cite{chen2024halc}, We employ the offline POPE (OPOPE) benchmark, using the F1-score as the evaluation metric for VH, where offline checks substitute the live interactions of POPE.
    As shown in Table~\ref{tab:res_pope}, we have several observations: 
    (1) \modelname consistently attains superior performance across most settings, surpassing both state-of-the-art methods, further validating its effectiveness;
    (2) \modelname effectively mitigates VH across three distinct LVLM architectures, showcasing its versatility and plug-and-play capability.   
    
\noindent\textbf{CHAIR Evaluation.}
    To evaluate our model's performance in open-ended caption generation with the CHAIR benchmark. Following HALC~\cite{chen2024halc}, we set `\textit{Please describe this image in detail.}' as the input prompt, as shown in Table~\ref{tab:chair}.
    From the results, we derive several detailed observations:
    (1) It can be observed that our method significantly outperforms existing methods for reducing VH. 
    (2)
    Our \modelname achieved the lowest VH rate at both the sentence and instance levels across three LVLM families while preserving recall, highlighting its superiority and generalizability in mitigating VH.
    
\noindent\textbf{MME Evaluation.}~
    Following~\cite{zhuang2025vasparse,chen2024halc},
    we adopt object-level subsets (``existence'' and ``count'') and attribute-level subsets ( ``position'' and ``color'') of MME benchmark~\cite{fu2023mme} to evaluate VH.
    As shown in Table~\ref{tab:res_mme_hallucination}, we can observe that:
    (1) Our \modelname can significantly reduce object and attribute hallucination, and achieve optimal VH mitigation performance.
    (2) DoLA and OPERA do not exhibit significant VH mitigation on the MME benchmark.
    This is because the MME evaluation is a binary classification task, requiring LVLMs to generate only a few tokens, which constrains methods reliant on longer sequences and special entity handling.
    
\begin{figure*}
  \centering
  \includegraphics[width=1\linewidth]{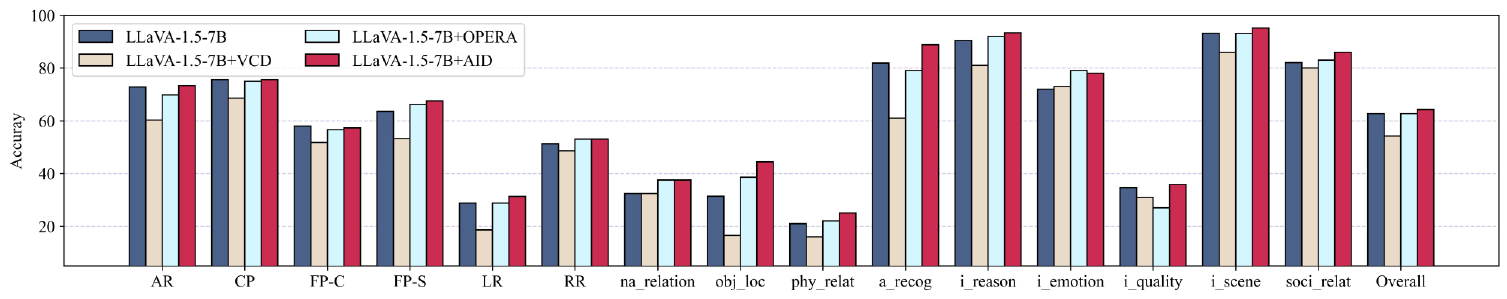}
    \vspace{-8mm}
   \caption{
   \textbf{Results on MMBench.} \modelname enhances comprehensive performance on diverse tasks.}
   \label{fig:mmbench}
   \vspace{-5mm}
\end{figure*}

\noindent \textbf{General-Purpose Evaluation.}
    To verify that our \modelname method does not compromise the model's generalization capability, we evaluated \modelname on multiple benchmarks, including MMBench, MM-Vet, and MME. 
    The results from MMBench (Fig. \ref{fig:mmbench}) show that AID is highly competitive compared to state-of-the-art (SOTA) methods, with detailed findings provided in Appendix. Notably, as shown in Tab. \ref{tab:mmvnetsub}, AID achieves significant overall performance improvements, particularly in OCR and spatial reasoning tasks, with an average increase of 6.7\%. 
    Finally, the model demonstrates superior generalization capabilities and enhanced hallucination mitigation when compared to VCD and OPERA method on MME (elaborated in Tab. \ref{tab:res_mme_hallucination}). Full experimental results are available in to Appendix for further reference.
  
\subsection{Ablation Studies}
\begin{figure}
      \centering
      \includegraphics[width=1\linewidth]{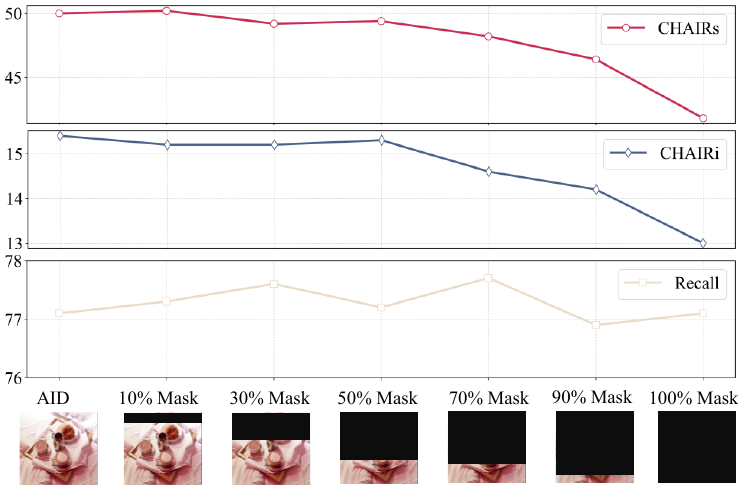}
       \caption{
        \textbf{Ablation Studies} of masking different proportions of visual areas in Attention Hijackers.
       }
       \label{fig:ablation_area_mask}
    \end{figure}

\noindent \textbf{Impact of Masking Different Visual Areas.}
    To evaluate the effect of masking Hijackers in various visual regions, we sequentially masked the Hijackers’ visual areas from top to bottom. The mask area was gradually increased to assess how AID's performance would respond, using CHAIR as the benchmark and LLaVA-1.5-7B as the backbone model. As shown in Fig.~\ref{fig:ablation_area_mask}, both CHAIR$_I$ ans CHAIR$_s$ scores increased as the mask moved downward. When the entire visual area was masked, performance improved significantly above the baseline, indicating that removing all visual information effectively isolates the impact of the Hijackers. This suggests that in self-interacting image regions, the lower area accumulates more visual information, with tokens in this region exerting a greater influence on the Hijackers.

    
       \begin{table}
        \begin{center}
        \small
        \resizebox{1.0 \linewidth}{!}{
        \begin{tabular}{llll}
        \toprule
        \bf Method   &\bf CHAIR$_S$ $\downarrow$ &\bf CHAIR$_I$ $\downarrow$ &\bf Recall $\uparrow$ \\ \midrule
         LLaVA-1.5 & 50.0 \basex{0.0} & 15.4 \basex{0.0} & 77.1 \basex{0.0} \\
         + AHD (Random-Mask)  & 50.8 \upbad{0.8} & 22.5 \upbad{7.1} & 77.3 \upbad{0.2}  \\
         + AHD (Mask-One)  & 44.8 \down{5.2} & 13.9 \down{1.5} & \bf{77.5} \down{0.4}  \\
         + AHD (Mask-Two) & 43.0 \down{7.0} & 13.2 \down{2.2} & \bf{77.5} \down{0.4}  \\
         + AHD (Mask-Three)  & 45.6 \down{4.4} & 13.9 \down{1.5} & \bf{77.5} \down{0.4}  \\
         + AHD (Mask-Two) + Linear Probing \cite{zhao2024first}  & 42.8 \down{7.2} & 13.5 \down{1.9} & 77.1 \basex{0.0}  \\
        \midrule
         \rowcolor{aliceblue} \textbf{+ AID (AHD + RD)}  & \bf{41.8} \down{8.2} & \bf{13.0} \down{2.4} & 77.1 \basex{0.0}  \\
        \bottomrule
        \end{tabular}}
        \end{center}
        \vspace{-1.em}
        \caption{\small \textbf{Ablation study for the designed components of AID.}}
        \label{tab:CHAIR_ablation}
        \vspace{-1.em}
        
    \end{table}
    \noindent \textbf{Effectiveness of Attention Hijacker Detection.}
    To evaluate each component of \modelname, we conduct ablation studies on Attention Hijackers Detection (AHD) and Re-Disentanglement (RD) using the CHAIR benchmark. As shown in Tab.~\ref{tab:CHAIR_ablation}, randomly masking instruction tokens (\textit{Random-Mask}) increase CHAIR$_s$ and CHAIR$_I$, likely due to unintentionally masking essential tokens, disrupting semantics, and increasing hallucinations. In contrast, AHD selectively masks Hijackers, reducing hallucinations by 5.2\% and 1.5\% on CHAIR$_s$ and CHAIR$_I$, demonstrating its effectiveness. Additionally, masking two Hijackers (\textit{Mask-Two}) yields optimal results, while masking three (\textit{Mask-Three}) offers less improvement. Thus, we empirically set the masked Hijacker count to 2.

    \begin{figure}[!t]
      \centering
      \includegraphics[width=1\linewidth]{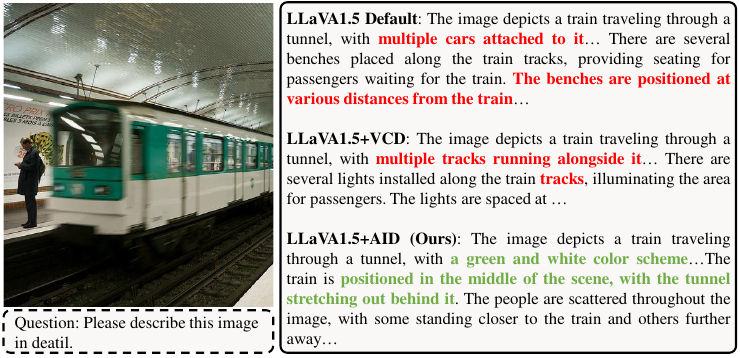}
       \caption{
        \textbf{Case Study.} AID effectively reduces hallucinations, outperforming VCD in visual scene understanding.
       }
       \label{fig:bad_case}
    \end{figure}
    
    \noindent \textbf{Effectieness of Re-Disentanglement.}
    To validate the effectiveness of Re-Disentanglement (RD) in addressing excessive masking, we compare it with the widely used Linear Probing approach~\cite{yu2024truth_linear,zhao2024first,DBLP:conf/acl/SuWAH00024,DBLP:conf/naacl/JiangQHFCMYZZ24}, which trains a linear model on hidden states to identify hallucinations while minimizing unnecessary operations. As shown in Tab.~\ref{tab:CHAIR_ablation}, \textit{AHD+Linear Probing} offers minimal improvement over \textit{AHD(Mask-Two)}, whereas \textit{AHD+RD} reduces hallucinations by 8.2\% and 2.4\% on CHAIR$_s$ and CHAIR$_I$, while maintaining recall. Unlike Linear Probing, RD requires no additional training, making it a more efficient solution.

        
    
    \subsection{Visualization Results}

To evaluate AID's effectiveness on image captioning, we compare it with LLaVA-1.5 and VCD. As shown in Fig. \ref{fig:bad_case}, LLaVA-1.5 generates hallucinated details, such as ``multiple cars attached" and ``benches positioned at various distances". In contrast, our AID accurately describes visual details like ``green and white color scheme" of the train and the people's positions relative to it, avoiding irrelevant information. This demonstrates that AID improves accuracy and reduces hallucinations, especially in complex scenes.

\section{Conclusion}
This paper presents a new perspective on hallucination mitigation, identifying it as interference from instruction tokens, or ``Attention Hijacking". We propose AID (\textbf{A}ttention H\textbf{I}jackers Detection and \textbf{D}isentanglement), a training-free strategy that detects and disentangles these Hijackers to restore proper attention. 
A potential negative impact is that AID focuses on improving image understanding and does not address factual accuracy issues. Future work, by integrating RAG, could help mitigate this limitation.

{
    \small
    \bibliographystyle{ieeenat_fullname}
    \bibliography{main}
}

\end{document}